\documentclass[letterpaper]{article} 
\usepackage{aaai24}  
\usepackage{times}  
\usepackage{helvet}  
\usepackage{courier}  
\usepackage[hyphens]{url}  
\usepackage{graphicx} 
\usepackage{natbib}  
\usepackage{caption} 
\usepackage{algorithm}
\usepackage{algorithmic}
\usepackage{booktabs}
\usepackage{multirow}
\usepackage{colortbl}
\usepackage{arydshln}
\usepackage{color}
\usepackage{xcolor}
\usepackage{bbding}
\definecolor{lightblue}{rgb}{0.93,0.95,1.0}
\usepackage{newfloat}
\usepackage{listings}
\DeclareCaptionStyle{ruled}{labelfont=normalfont,labelsep=colon,strut=off} 
\floatstyle{ruled}
\newfloat{listing}{tb}{lst}{}
\floatname{listing}{Listing}
\nocopyright 

\title{FlexKBQA: A Flexible LLM-Powered Framework for Few-Shot Knowledge Base Question Answering}
\author{
    Zhenyu Li\textsuperscript{\rm 1}\equalcontrib,
    Sunqi Fan\textsuperscript{\rm 1}\equalcontrib,
    Yu Gu\textsuperscript{\rm 2},
    Xiuxing Li\textsuperscript{\rm 3}\thanks{Corresponding author},
    Zhichao Duan\textsuperscript{\rm 1},\\
    Bowen Dong\textsuperscript{\rm 1},
    Ning Liu\textsuperscript{\rm 4},
    Jianyong Wang\textsuperscript{\rm 1\dag}
}
\affiliations{
    \textsuperscript{\rm 1}Tsinghua University \quad
    \textsuperscript{\rm 2}The Ohio State University\\
    \textsuperscript{\rm 3}University of Chinese Academy of Sciences	 \quad
    \textsuperscript{\rm 4} Shandong University \\

}

\usepackage{bibentry}

\begin{document}

\maketitle

\begin{abstract}
Knowledge base question answering (KBQA) is a critical yet challenging task due to the vast number of entities within knowledge bases and the diversity of natural language questions posed by users. 
Unfortunately, the performance of most KBQA models tends to decline significantly in real-world scenarios where high-quality annotated data is insufficient.
To mitigate the burden associated with manual annotation, we introduce {\it \textbf{FlexKBQA}} by utilizing Large Language Models (LLMs) as program translators for addressing the challenges inherent in the few-shot KBQA task. 
Specifically, {\it \textbf{FlexKBQA}} leverages automated algorithms to sample diverse programs, such as SPARQL queries, from the knowledge base, which are subsequently converted into natural language questions via LLMs. This synthetic dataset facilitates training a specialized lightweight model for the KB.
Additionally, to reduce the barriers of distribution shift between synthetic data and real user questions, {\it \textbf{FlexKBQA}} introduces an execution-guided self-training method to iterative leverage unlabeled user questions. Furthermore, we explore harnessing the inherent reasoning capability of LLMs to enhance the entire framework. Consequently, {\it \textbf{FlexKBQA}} delivers substantial flexibility, encompassing data annotation, deployment, and being domain agnostic.
Through extensive experiments on GrailQA, WebQSP, and KQA Pro, we observe that under the few-shot even the more challenging zero-shot scenarios, {\it \textbf{FlexKBQA}} achieves impressive results with a few annotations, surpassing all previous baselines and even approaching the performance of supervised models, achieving a remarkable 93\% performance relative to the fully-supervised models. We posit that {\it \textbf{FlexKBQA}} represents a significant advancement towards exploring better integration of large and lightweight models. The source code and pertinent documentation are readily accessible on established open-source repositories~\footnote{\url{https://github.com/leezythu/FlexKBQA}}.

\end{abstract}

\section{Introduction}

\begin{figure}[t]
\centering
\includegraphics[width=0.9\columnwidth]{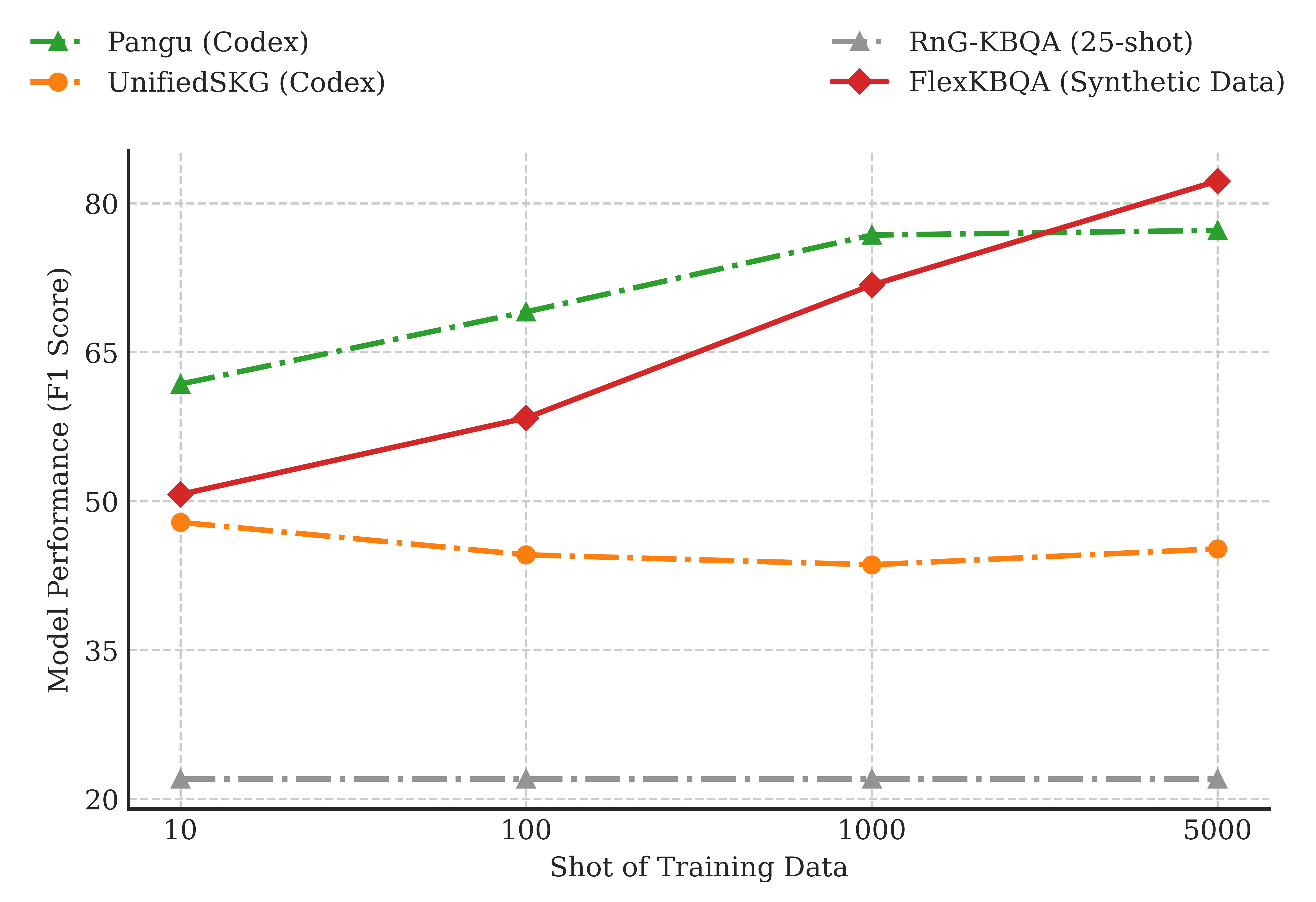} 
\caption{An analytical experiment on 500 random samples of GrailQA dev set (oracle entity linking). FlexKBQA's performance exhibits a consistently upward trend with the increasing synthetic data size, surpassing all in-context learning models with limited window length. Since our synthetic data are generated based on 25-shot real data, we also depict the performance of our underlying model (RnG-KBQA) trained by 25-shot real data as a baseline.}
\label{fig1}
\end{figure}

Knowledge base question answering (KBQA) plays a crucial role in leveraging the substantial knowledge stored in knowledge bases and making it accessible to users \cite{berant2013semantic,yih2016value,gu2022knowledge}. 
With the ever-growing size and complexity of  knowledge bases (KBs), effective KBQA systems are becoming increasingly important.
Many current KBQA systems depend on supervised training using large sets of manually annotated data. 
This approach brings about significant labor challenges for two main reasons: 
(1) The substantial size and comprehensive scope of the KB, coupled with its ever-evolving content, lead to a vast sample space~\cite{berant2013semantic,gu2021beyond}. 
This complicates the process of collecting a representative and exhaustive set of training data.
(2) The heterogeneity in KB schema and design protocols, combined with differences in the query languages, such as SPARQL~\cite{su2016generating}, S-expression~\cite{gu2021beyond}, and KoPL~\cite{cao2022kqa}, requires tailored training processes for each KB. 
This intensifies the labor involved in developing and adapting models for each distinct KB.
Considering these complexities, there emerges an imperative to forge \emph{a flexible framework that can efficiently build adaptable KBQA models for different KBs and query languages, utilizing only a limited set of annotated examples}.

The emergence of large language models (LLMs) in recent times, such as Codex \cite{chen2021evaluating} and GPT-4 \cite{openai2023gpt4}, suggests promising avenues for constructing such frameworks.
LLMs have showcased versatility across diverse tasks and remarkable generalization with minimal demonstrations~\cite{wei2022chain,li2023fewshot}, as exemplified in ~\citet{cheng2023binding}, where Codex outperforms previous fine-tuning methods with only a handful of examples. 
Such feats hint at LLMs' potential in KBQA.
However, research on LLMs' utility for few-shot KBQA remains scant.
Initial explorations by \citet{li2023fewshot} and \citet{gu2023dont} employ in-context learning, allowing LLMs to transform questions into programs using minimal demonstrations. 
Nonetheless, this paradigm possess several limitations. 
First, LLMs inherently scuffle with constraints such as limited context window and considerable inference overhead. 
Furthermore, LLMs, despite their generalization capacities, face challenges in addressing the intricacies embedded within domain-specific KBs when not fine-tuned, leading to inaccuracies when generating target entities and relations~\cite{li2023fewshot}.
Additionally, without fine-tuning, the benefits of in-context learning diminish as more training data is introduced~\cite{gu2023dont}.

In light of these challenges, we present FlexKBQA (see Figure~\ref{fig2}), a flexible KBQA framework harnesses the advanced generative abilities of LLMs to produce synthetic data, which in turn aids in the training of lightweight models.
Specifically, FlexKBQA first leverages collected templates of structured queries to sample a substantial number of programs (i.e., S-expressions) from the KB. 
These programs are then converted into coherent natural language questions using LLMs, ensuring both the accuracy of the programs and the fluency of the generated questions. 
Upon generating these program-question pairs, they serve as valuable resources for the fine-tuning of lightweight models.
However, there could be a potential distribution shift between synthetic data and real user queries. 
To bridge this gap, we further introduce the execution-guided self-training (EGST) method, 
which employs the fine-tuned lightweight model to proactively annotate real user queries over the KB.
These annotated queries subsequently serve as valuable training data, enabling self-improvement.
FlexKBQA addresses the inherent constraints of the context window of LLMs, enabling the KBQA model to exhibit consistent improvement as more training data is incorporated (Figure~\ref{fig1}). Additionally, by delegating the inference process to lightweight models rather than relying on the LLM directly, our approach ensures enhanced efficiency in inference.


We conduct extensive experiments on three representative datasets - GrailQA, WebQSP, and KQA Pro. FlexKBQA outperforms existing models by a large margin in the few-shot setting.
Remarkably, with a mere 25 labeled examples, FlexKBQA surpasses the performance of all previous methods utilizing 100 shots on GrailQA. 
Furthermore, FlexKBQA's results are comparable to several fully supervised models on both GrailQA and WebQSP datasets.

The main contributions of the paper can be highlighted as follows:
\begin{itemize}
    \item We present an efficient and flexible KBQA framework, capitalizing on the notable generative capabilities of LLMs.
    \item We introduce the execution-guided self-training strategy to address the distribution shift challenge by facilitating interaction between heterogeneous information and leveraging the inherent reasoning ability of LLM.
    \item Experimental results demonstrate that FlexKBQA significantly outperforms all baselines on real-world datasets under different settings.
    \item To the best of our knowledge, our work represents the inaugural endeavor in exploring the zero-shot KBQA.
    \end{itemize}

\section{Related Work}
\subsection{Knoweldge Base Question Answering}
One prominent line of research on KBQA involves semantic parsing, where natural language questions are mapped to formal representations, such as SPARQL queries \cite{lan2021survey,cao2022program}, enabling efficient querying of structured knowledge bases. Recently, researchers have proposed using step-wise discriminative approaches instead of directly generating programs in an autoregressive fashion to effectively limit the massive search space and reduce errors, leading to state-of-the-art results \cite{gu2022arcaneqa, gu2023dont}. However, these models rely on a substantial number of annotated training samples from the KB. Developing a usable QA system with only a limited amount of annotated data from the KB is of great practical significance.
\subsection{Few-Shot Language Understanding with LLMs}


Researchers have been dedicated to studying the language understanding capabilities in the zero-shot or few-shot setting \cite{gao2019fewrel,li2022effective,li2023toward}. In recent years, LLMs have demonstrated strong few-shot learning abilities across various tasks, such as question answering \cite{cheng2023binding}, code generation \cite{ni2023lever}, and embodied agents \cite{singh2023progprompt}. Recently, Pangu \cite{gu2023dont} and KB-BINDER \cite{li2023fewshot} have pioneered the exploration of the few-shot setting on the KBQA task. They conduct experiments under the in-context learning paradigm, providing a few question-program pairs and allowing the large language model to leverage its completion capability to make new predictions. Despite the remarkable generalization ability of large language models, they are still limited by the inherent window size of in-context learning, which constrains their performance. Additionally, large models come with drawbacks in terms of cost and security. Therefore, in this paper, we explore a hybrid approach that combines large and lightweight models, aiming to achieve improved performance while being more deployable.

\subsection{Combination of LLMs and light-weight models}
Although LLMs have demonstrated strong capabilities across various tasks, they are computationally expensive, slow during inference, and difficult to deploy. Numerous research endeavors have been focused on exploring the interaction and complementary aspects between large-scale models and lightweight models, e.g., distilling the knowledge from large-scale models to lightweight models by matching the output distribution \cite{hinton2015distilling}. Despite the traditional knowledge distillation, researchers propose a new paradigm called teaching via data (TvD), in which they use an LLM-based “teacher” model to generate synthetic data for a specific task, then use the data to fine-tune a smaller “student” model \cite{schick2021generating,rosenbaum2022clasp, rosenbaum2022linguist, meng2022generating,ye2022zerogen}.
The types of synthetic data generated by LLM are diverse. SYMGEN \cite{ye2023generating} first proposes using LLM to generate symbolic language with execution-based verification. Our work adopts the teaching via data paradigm but we are the first to utilize LLM as program translator, thereby addressing the challenge of annotating KBQA training data.
\begin{figure*}[ht]
\centering
\includegraphics[width=1.9\columnwidth]{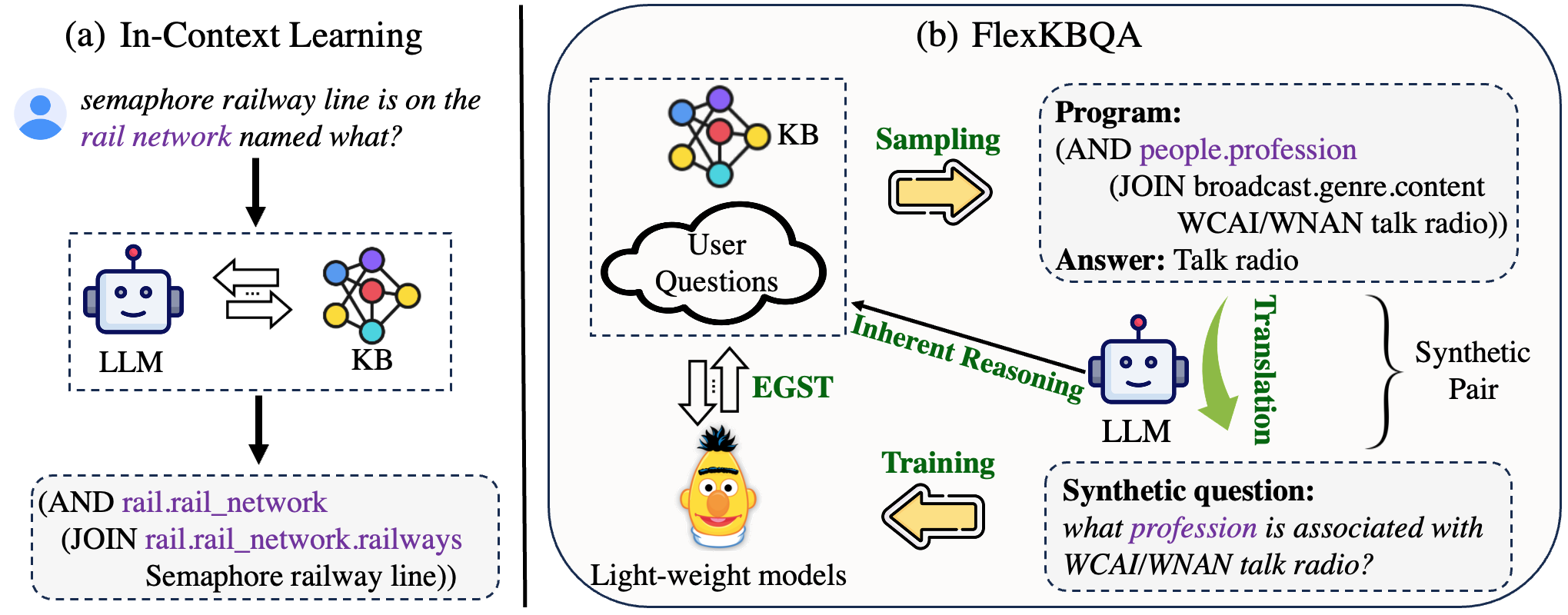} 
\caption{A comparison between FlexKBQA and prior methods. (a): 
 Prior approaches enable LLMs to directly ground the question to the knowledge base through in-context learning capabilities. (b): An illustration of FlexKBQA's innovative design: (1) Automatic Program Sampling module generates diverse and executable programs. (2) Low-Resource Program Translation module synthesizes high-quality data pairs. (3) Execution-Guided Self-Training module addresses distribution shift. (4) Inherent Reasoning module boosts the pipeline by leveraging inherent knowledge within LLMs.}
\label{fig2}
\end{figure*}

\section{Methodology}
\subsection{Preliminaries}
A knowledge base can be formally represented as $\mathcal{K} \in \mathcal{E} \times \mathcal{R} \times (\mathcal{E} \cup \mathcal{L} \cup \mathcal{C})$, where $\mathcal{E}$ represents the set of entities, $\mathcal{R}$ denotes the set of relations between entities, and $\mathcal{C}$ denotes the set of classes. The primary goal of KBQA is to answer user questions by leveraging the facts stored in the knowledge base.
For accuracy and interpretability, most existing work adopt a semantic parsing framework, transforming natural language questions into programs, such as S-expression or SPARQL. These programs can be executed on the knowledge base to retrieve the final answer.

Most models utilize a framework comprising a ranking model, a generation model, or a combination of both. The ranking model is usually optimized as follows:
\begin{equation}
    \mathcal{L}_{ranker} = - \frac{ e^{s(q,p)} }{e^{s(q,p)}+\sum_{p_i\in P \wedge p_i\neq 
 p}e^{s(q,p_i)}}
\end{equation}
where $p$ is the target program, and $s(q, p_i)$ is the similarity score between the question $q$ and each program $p_i$ from a candidate set $P$. The score is derived from a lightweight model like BERT \cite{devlin-etal-2019-bert}. On the other hand, a generation model employs advanced pre-trained generative models, such as T5, to generate the target program:
\begin{equation}
    \mathcal{L}_{gen} = -\sum_{t=1}^{n}log(probability(p_t|p_{<t};q;r))
\end{equation}
where $n$ is the length of $p$, and $r$ represents additional information like ranking results.
\subsection{Automatic Program Sampling}
The objective of program sampling is to generate valid (executable on KB) programs. We break this process into two crucial steps: template collection and step-wise grounding.


\textbf{Template collection}
Taking a SPARQL query as an example, if we replace the entities and relations within it with variables (e.g., $ent0, rel0, ent1$), the resulting structural form is referred to as a ``template". By employing automated algorithms, we can generate a diverse set of templates to cover various question types that users may encounter in real-world scenarios. Compared to annotating programs for a large number of questions, the cost of collecting a handful of templates is negligible.

\textbf{Step-wise grounding}
After obtaining a collection of diverse program templates, the next step is to perform entity and relation grounding. Leveraging the querying mechanism of SPARQL, we can directly treat the variables as query objects. However, directly executing the query with multiple variables can result in long execution times or errors, particularly when dealing with large-scale knowledge bases. To address this challenge, we introduce a ``step-wise grounding" approach, where we iteratively determine the values of variables. Through empirical assessment, we note that this strategy efficiently narrows down the search space while maintaining diversity, ultimately enabling the derivation of a substantial number of programs. We give a specific example in Appendix A to illustrate this process.

\subsection{Low-Resource Program Translation}
As illustrated in Figure \ref{fig2}, conventional approaches often utilized language models to directly convert questions into programs via in-context learning. However, given that LLMs are primarily trained on natural language and lack specific training on programs, we argue that translating programs into natural language questions is a more efficient and intuitive approach, especially under the low-resource setting.

In this paper, we consider the LLM as a program translator that converts a program $p_i^s$ into its corresponding natural language question $q_i^s$.
The prompt is composed of (1) an instruction $Inst$ to guide the model in transforming programs into natural language questions. (2) a few seed pairs of golden programs and questions $\{(p_1^f,q_1^f),...,(p_N^f,q_N^f)\}$ as demonstrations, where $N$ denotes a $N$-shot setting, and $N=0$ represents the zero-shot setting. Note that when constructing seed pairs, we should prioritize selecting diverse programs to cover a wide range of question types. 
Overall, the translation process can be formulated as:
\begin{equation}
   q_i^s \leftarrow Translator(Inst;
   (p_1^f,q_1^f),...,(p_N^f,q_N^f);p_i^s)
\end{equation}
For detailed prompts, please refer to Appendix B.
\subsection{Execution-Guided Self Training}
After generating a synthetic dataset using the aforementioned steps for training lightweight models, a potential issue still remains – the distribution discrepancy between synthetic and real-world questions. This discrepancy primarily arises from the limited overlap of entities and relations. Existing literature 
emphasizes that distribution shifts can have a substantial impact on the model's performance. For instance, a model trained on the GrailQA dataset might achieve only around 65\% performance compared to being trained directly on the WebQSP \cite{gu2021beyond}. Therefore, in this paper, we propose using execution-guided self-training (\textbf{EGST}) to address this issue. Taking into account the ease of collecting user questions in real-world scenarios, we apply the self-training method to harness the information from these unlabeled user questions. Moreover, we introduce the utilization of execution results from structured queries as feedback information, substantially enhancing the purity of the training samples.

Specifically, the entire training process follows a teacher-student iterative training approach shown in Algorithm \ref{alg1}. Firstly, we train a teacher model using the synthetic data generated by the LLM. Subsequently, in each iteration, the teacher model generates pseudo programs for unlabeled real user questions, which undergo an execution-guided filtering mechanism to remove noisy data and obtain a cleaner training dataset.  This filtered dataset, alongside the synthetic data, is employed to train the student model, which serves as the teacher model for the subsequent iteration. This iterative process continues until convergence is achieved.

For the execution-guided filtering mechanism, we employ the following filtering rules during each iteration (more details and explorations in Appendix C and H):
\begin{itemize}
    \item Error Filtering: we remove those data pairs where the predicted SPARQL queries either result in execution errors or fail to retrieve answers.
    \item Semantic Filtering: we employ an off-the-shelf pre-trained sentence-transformer {\cite{reimers2019sentencebert}} to calculate the semantic textual similarity between natural language questions and the relations in predicted programs, and filter out those data pairs with low similarity.
    \item Inherent Reasoning Filtering: We exclude samples whose pseudo answers do not align with the outcomes of Inherent Reasoning. Further details are provided in the next section.
\end{itemize}


\begin{algorithm}[tb] 
  \caption{Execution-Guided Self Training Procedure}
  \label{alg1}
\textbf{Input}: Unlabeled user questions $D^u = \{(q_i^u)\}$ ;\\ A few labeled question-program pairs $D^f=\{(q_i^f,p_i^f)\}$;\\  Synthetic pairs from LLM $D^s=\{(q_i^s,p_i^s)\}$;\\Initial model parameters $\theta_{ini}$\\
\textbf{Output}: Model parameter $\theta_{final}$ 
  \begin{algorithmic}[1]
 \STATE  Fine-tune teacher model $\theta_{tea}$ on synthetic data and small labeled data $D^s\cup D^f$ from $\theta_{ini}$\\
 \STATE \textbf{while} not converged \textbf{do}  
\STATE \quad \quad //Generate pseudo-label for user questions
\\ \quad \quad $p_i^u \leftarrow f(q_i^u;\theta_{tea})$
    \STATE \quad \quad //Execution-guided filtering
   \\ \quad \quad  $\{(q_i^{uf},p_i^{uf})\} \leftarrow Filter(KB,\{(q_i^u,p_i^u)\}$)
    \STATE \quad \quad //Fine-tune student model $\theta_{stu}$ on  all data 
    \\ \quad \quad  $D=\{(q_i^s,p_i^s)\}\cup \{(q_i^f,p_i^f)\}\cup \{(q_i^{uf},p_i^{uf})\}$
    \STATE \quad \quad //Update the teacher model \\ \quad \quad $\theta_{tea} \leftarrow \theta_{stu}$
\STATE \textbf{end while}
\STATE $\theta_{final}\leftarrow\theta_{tea}$
  \end{algorithmic}

\end{algorithm}
\vspace{-1em} 

\subsection{Inherent Reasoning Augmentation}
A widely accepted view is that LLMs encode knowledge in their parameters, allowing them to answer complex questions and directly respond to user queries. We refer this approach as Inherent Reasoning (\textbf{IR}). However, compared to transforming user questions into executable programs on KBs, Inherent Reasoning lacks interpretability and could exhibit limited effectiveness when dealing with domain-specific KBs. Thus, in this paper, we employ Inherent Reasoning as a data augmentation technique. It primarily operates in two stages. \textbf{(1)} In the execution-guided self-training stage, after annotating pseudo programs and obtaining answers for real user questions, we select samples where the answers align with those generated through Inherent Reasoning. By combining Inherent Reasoning with execution-guided filtering, we can obtain a training dataset with reduced noise. \textbf{(2)} Inherent Reasoning can serve as a complementary approach to semantic parsing. When the semantic parsing method fails to retrieve an answer (e.g. cannot enumerate ranking candidates or the predicted programs result in execution errors), we resort to using results from Inherent Reasoning as the final answers for the initially unanswerable questions. We believe that the integration of Inherent Reasoning and the semantic parsing approach holds significant value in the development of more accurate and interpretable KBQA systems. Please refer Appendix B for more details.

In summary, with the aforementioned core components, FlexKBQA can achieve the following dimensions of flexibility:
(1) Data Efficient: FlexKBQA requires only a small number of question-program pairs as prompts.
(2) Domain-Agnostic: FlexKBQA is applicable to diverse knowledge bases, and alleviates common distribution shifting issues.
(3) Deployable: Utilizing a lightweight model allows for cost-effective deployment compared to closed-source LLMs and enables seamless integration of domain-specific knowledge through fine-tuning.
\section{Experimental Setup}

\subsection{Datasets}
We conduct experiments on datasets from different KB and of diverse program types (statistics in Table \ref{datasets}).\newline
\textbf{GrailQA} \cite{gu2021beyond} dataset is a large-scale dataset for knowledge base question answering that contains 64,331 question-logical form pairs. It has a broad coverage and many unique canonical logical forms. GrailQA introduces three levels of generalization for KBQA:  \textit{i.i.d.}, \textit{compositional} and \textit{zero-shot}. This dataset has attracted substantial research interest in recent years.
 \newline
 \textbf{WebQSP} \cite{yih2016value} is another widely recognized KBQA dataset. The questions in this dataset are extracted from Google query logs, making them reflective of user preferences in real-world scenarios. It also offers SPARQL annotations that can be executed directly on Freebase. The main objective of this dataset is to evaluate the generalization capability in an \textit{i.i.d.} setting, as the training and testing data share common entities and relations.\newline
\textbf{KQA Pro} \cite{cao2022kqa} consists of around 120,000 diverse natural language questions that require various reasoning capabilities, such as multi-hop inference, attribute comparison, and set operations. KQA Pro is constructed based on a sub-knowledge base from Wikidata and does not assume an entity linking stage. Therefore, it requires models to memorize entities and relations, making it a more strong and challenging \textit{i.i.d} task.

\begin{table}[t]
  \centering
    \begingroup
  \renewcommand{\arraystretch}{0.6}
  \begin{tabular}{ccccc} \toprule
  Dataset &  \ KB & \ Questions  & \ Program Type\\ \midrule
\multirow{2}{*}{GrailQA}  & \multirow{2}{*}{Freebase}  & \multirow{2}{*}{64,331}  & \multirow{2}{*}{S-expression}\\ 
&&&&\\ \midrule

\multirow{2}{*}{WebQSP}    & \multirow{2}{*}{Freebase} & \multirow{2}{*}{4,737} & \multirow{2}{*}{SPARQL} \\ 
&&& &\\
\midrule

\multirow{2}{*}{KQA Pro}    & \multirow{2}{*}{Wikidata} & \multirow{2}{*}{117,970} & \multirow{2}{*}{SPARQL} \\ 
&&&&\\ 
 \bottomrule
  \end{tabular}
  \caption{Dataset Statistics}
  \label{datasets}
  \endgroup
\end{table}

\subsection{Underlying Model and Baselines}
Note that FlexKBQA is model-agnostic. Considering performance and reproducibility, we choose a well performing RnG-KBQA~\cite{ye2022rng} as the underlying model for GrailQA and WebQSP. For KQA Pro, we select the BART-SPARQL~\cite{cao2022kqa} model. 

For baselines, we mainly evaluate FlexKBQA against Pangu \cite{gu2023dont} and KB-BINDER \cite{li2023fewshot}. Both of them leverage the potent large language model Codex for in-context learning. Note that because the few-shot KBQA is a relatively novel task, our baseline choices are limited to these two methods. Since there are no available experimental results for Pangu and KB-BINDER on KQA Pro, we also re-implement an in-context learning model called LLM-ICL as an alternative for evaluation. We also provide the supervised results of several representative models for comparison. We provide detailed descriptions of these models in Appendix D.

\subsection{Implementation Details}
In experiments, we consider the original training dataset as the unlabeled real user questions.
During the ``step-wise grounding" stage, we not only employ random sampling but also gather the set of entities present in unlabeled user questions. These entities are then used as ``topic entities" for constructing SPARQL queries. After removing duplicates, we obtain 6,184 synthetic pairs on Freebase and 5,017 on Wikidata. 
For a fair comparison, we use the same off-the-shelf entity linkers as Pangu \cite{gu2023dont}. And the LLM we utilize for program translation is gpt-3.5-turbo\footnote{https://platform.openai.com/docs/models/gpt-3-5}.
More details can be found in Appendix E.

\begin{table*}[ht]
  \centering
  \scalebox{0.95}{
  \begin{tabular}{c|l|ccllcccccc}
  \toprule
   \multicolumn{2}{c}{}  &  \multicolumn{2}{c}{\textbf{Overall}}&\multicolumn{2}{c}{\textbf{I.I.D.}}&\multicolumn{2}{c}{\textbf{Compositional}}&\multicolumn{2}{c}{\textbf{Zero-shot}}&\multicolumn{2}{c}{\textbf{Dev Overall}}\\
  &\textbf{Model} & \textbf{EM} & \textbf{F1}& \textbf{EM} & \textbf{F1}& \textbf{EM} & \textbf{F1}& \textbf{EM} & \textbf{F1}& \textbf{EM} & \textbf{F1}\\ 
\midrule
\multirow{6}{*}{Supervised}& QGG  \cite{c:0}   &-&36.7&-&40.5&-&33.0&-&36.6&-&-\\
&BERT+Ranking \cite{gu2021beyond}& 50.6 &58.0&59.9&67.0&45.5&53.9&48.6&55.7&-&- \\
&ReTraCk  \cite{chen2021retrack}& 58.1&65.3&84.4&87.5&61.5&70.9&44.6&52.5&-&-\\ 
&RnG-KBQA  \cite{ye2022rng}&68.8&74.4&86.2&89.0&63.8&71.2&63.0&69.2&71.4&76.8\\ 
&ArcaneQA  \cite{gu2022arcaneqa}&63.8&73.7&85.6&88.9&65.8&75.3&52.9&66.0&69.5&76.9\\ 
& DecAF  \cite{yu2023decaf}& 68.4&78.7&84.8&89.9&73.4&81.8&58.6&72.3&-&81.4 \\
\midrule 
\multirow{2}{*}{\shortstack{Few-Shot \\ (100 shots)}}
& KB-BINDER \cite{li2023fewshot} & 50.6&56.0&-&-&-&-&-&-&-&-\\
& Pangu \cite{gu2023dont}&53.3&62.7&54.7&62.9& 54.5 &63.7& 52.3 &62.2&-&-\\ 
\cdashline{1-12}[1pt/1pt]
\multirow{4}{*}{\shortstack{Few-Shot \\ (25 shots)}}
&Fine-Tuning  & 16.6&21.3&19.0&24.8&17.3&21.8&15.2&19.4&16.4&21.6\\

&\cellcolor{lightblue}{\textbf{FlexKBQA}} & \cellcolor{lightblue}{\textbf{62.8}}& \cellcolor{lightblue}{\textbf{69.4}} & \cellcolor{lightblue}{\textbf{71.3}} & \cellcolor{lightblue}{\textbf{75.8}} & \cellcolor{lightblue}{\textbf{59.1}} & \cellcolor{lightblue}{\textbf{65.4}} & \cellcolor{lightblue}{\textbf{60.6}} & \cellcolor{lightblue}{\textbf{68.3}} & \cellcolor{lightblue}{\textbf{65.5}} & \cellcolor{lightblue}{\textbf{71.1}} \\ 
&\textbf{\quad \textit{-w/o} IR} & \cellcolor{lightblue}{\textbf{62.8}} & 68.0 & \cellcolor{lightblue}{\textbf{71.3}} & 75.3 & \cellcolor{lightblue}{\textbf{59.1}} & 64.1 & \cellcolor{lightblue}{\textbf{60.6}} & 66.4 & \cellcolor{lightblue}{\textbf{65.5}} & 70.6\\

&\textbf{\quad \textit{-w/o} EGST } & 52.4 & 57.7 & 56.9 & 61.8 & 49.4 
& 54.4 & 51.7 & 57.3 & 57.0 & 62.1\\
\midrule
\multirow{3}{*}{Zero-Shot}
&\textbf{FlexKBQA} & 61.9 & 68.9 & 72.1 & 77.0 & 58.4 & 65.2 & 58.9 & 66.9 & 64.6 & 70.3 \\ 
&\textbf{\quad \textit{-w/o} IR} & 61.9 & 67.6 & 72.1 & 76.5 & 58.4 & 64.1 & 58.9 & 65.0 & 64.6 & 69.8\\

&\textbf{\quad \textit{-w/o} EGST } & 51.9 & 57.5 & 56.1 & 60.5 & 49.6 
& 53.7 & 52.9 & 57.8 & 56.4 & 61.2\\
  \bottomrule
  \end{tabular}
  }
     \caption{Results on GrailQA}
  \label{results on grailqa}
\end{table*}


\begin{table}[tb]
  \centering
  \scalebox{0.95}{
  \begin{tabular}{c|l|c}
  \toprule
 & \textbf{Model} & \textbf{F1}\\
\midrule
\multirow{6}{*}{Supervised}& QGG  \cite{c:0}   &74.0\\
&ReTraCk  \cite{chen2021retrack}& 71.0\\ 
&CBR  \cite{das2021case}&72.8\\ 
&Program Transfer  \cite{cao2022program}&76.5\\ 
&RnG-KBQA  \cite{ye2022rng}&75.6\\ 
& DecAF  \cite{yu2023decaf}& 78.8 \\
\midrule 
\multirow{6}{*}{\shortstack{Few-Shot\\(100 shots)}}
 & Fine-Tuning & 25.6 \\
& KB-BINDER \cite{li2023fewshot} & 53.2\\
& Pangu \cite{gu2023dont}&54.5\\
&   \cellcolor{lightblue}{\textbf{FlexKBQA} }&\cellcolor{lightblue}{\textbf{60.6}}\\
&   \textbf{\quad \textit{-w/o} IR} & 58.2\\
& \textbf{\quad  \textit{-w/o} EGST}& 51.1\\
\midrule
\multirow{3}{*}{Zero-Shot}
&   \textbf{FlexKBQA} & 46.2 \\
&   \textbf{\quad \textit{-w/o} IR} & 45.7 \\
& \textbf{\quad  \textit{-w/o} EGST}& 33.4 \\

  \bottomrule
  \end{tabular}}
\caption{Results on WebQSP}
  \label{results on webqsp}

\end{table}

\begin{table}[tb]
  \centering
  \scalebox{0.9}{
  \begin{tabular}{c|l|c}
  \toprule
 & \textbf{Model} &\textbf{Accuracy} \\
\midrule
\multirow{4}{*}{Supervised}&
EmbedKGQA (Saxena et al. 2020) & 28.36 \\
&RGCN  \cite{schlichtkrull2018modeling}   &35.07\\
&RNN SPARQL \cite{cao2022kqa} &41.98\\
&BART+SPARQL  \cite{cao2022kqa}&89.68 \\ 
\midrule 
\multirow{5}{*}{\shortstack{Few-Shot\\(100 shots)}}& 
Fine-Tuning &22.45 \\ 
& LLM-ICL & 31.75 \\
&\cellcolor{lightblue}{\textbf{FlexKBQA } }&\cellcolor{lightblue}{\textbf{46.83}} \\
&\textbf{\quad \textit{-w/o} IR }  &33.32 \\
&\textbf{\quad \textit{-w/o} EGST} &23.10  \\
\midrule
\multirow{3}{*}{Zero-Shot}& 
\textbf{FlexKBQA  }  &33.28 \\
&\textbf{\quad \textit{-w/o} IR }  &11.11 \\
&\textbf{\quad \textit{-w/o} EGST} &8.24  \\
  \bottomrule
  \end{tabular}
  }
 \caption{Results on KQA Pro}
  \label{results on kqapro}
\end{table}

\section{Results}

\subsection{Main Results}
The results are presented in Table \ref{results on grailqa}, \ref{results on webqsp}  and \ref{results on kqapro}.
Compared to other baselines that also target the few-shot KBQA scenario, FlexKBQA demonstrates significant superiority. On the test set of GrailQA , it achieved an impressive Exact Match (EM) score of 62.8 and an F1 score of 69.4 with only 25 annotated samples, outperforming the previous state-of-the-art model Pangu by a significant margin of 6.7 points in terms of F1 score, despite Pangu utilizing more shots. 
Surprisingly, FlexKBQA also surpasses several supervised models, such as ReTraCk, which demand training with tens of thousands of samples. Since we use  RnG-KBQA model as the underlying model of FlexKBQA, it's interesting to note that FlexKBQA attains a remarkable 93\% performance relative to the fully-supervised RnG-KBQA model. 

On WebQSP and KQA Pro datasets, FlexKBQA exhibits a similar trend of remarkable superiority over in-context learning methods. Specifically, when considering the F1 score, FlexKBQA achieves a significant 6.1-point improvement over Pangu on WebQSP under the 100-shot setting. It is worth noting that FlexKBQA's performance on the KQA Pro dataset shows a gap compared to the best-performing model. This difference can be attributed to the absence of an entity linking stage in KQA Pro. As a result, if the relations or entities in the test set were not present during training, FlexKBQA is more prone to producing semantic correct but unexecutable programs.

We also conduct experiments under a more challenging zero-shot setting, assuming no available annotated samples at all. This scenario has been rarely explored in previous research. We observe that, although the performance is behind the level achieved in the few-shot scenario, results on GrailQA show significant potential of our method. And we believe that this direction is worth further exploration.

\begin{table*}[htbp]
    \centering
\scalebox{0.8}{
    \begin{tabular}{cc}
    \toprule
\textbf{Question \uppercase\expandafter{\romannumeral1} } & \textbf{\it{What type of art leonardo da vinci do?}}  \\ \midrule
Pangu & (JOIN (R \textcolor{red}{visual\_art.visual\_artist.associated\_periods\_or\_movements}) m.04lg6)\,({\color{red}{\XSolidBrush}})
\\ 
 FlexKBQA & 
 (JOIN (R \textcolor[RGB]{76,187,23}{visual\_art.visual\_artist.art\_forms}) m.04lg6)
 \, ({\color[RGB]{76,187,23}{\Checkmark}})\\
 \midrule
\multirow{2}{*}{Synthetic Data Pair} & \it{What type of art did andy warhol create?} \\
& (JOIN (R \textcolor[RGB]{76,187,23}{visual\_art.visual\_artist.art\_forms}) m.0kc6) \\
\midrule

\textbf{Question \uppercase\expandafter{\romannumeral2} } & \textbf{\it{What airport do you fly into to get to destin fl?}}  \\ \midrule
Pangu & (JOIN (R \textcolor{red}{travel.travel\_destination.tourist\_attractions}) m.0rp8x)\,({\color{red}{\XSolidBrush}})\\ 
FlexKBQA \textit{w/o} EGST & (JOIN (R \textcolor{red}{travel.travel\_destination.how\_to\_get\_here}) m.0rp8x)\,({\color{red}{\XSolidBrush}})\\
FlexKBQA & (JOIN (R \textcolor[RGB]{76,187,23}{location.location.nearby\_airports}) m.0rp8x)\,({\color[RGB]{76,187,23}{\Checkmark}})\\
 \midrule
\multirow{2}{*}{Pseudo Labeled Pair } & \it{What airport is closer to downtown houston?}\\
& (JOIN (R \textcolor[RGB]{76,187,23}{location.location.nearby\_airports}) m.03l2n) \,({\color[RGB]{76,187,23}{\Checkmark}}) \\
\bottomrule
 
    \end{tabular}
    }
   \caption{Two typical questions from the test set of WebQSP that FlexKBQA succeeds while Pangu fails. The incorrect relations are marked as red, while the correct relations are marked as green.}
     \label{table5}
\end{table*}

\subsection{EGST and IR}

According to Tables \ref{results on grailqa}, \ref{results on webqsp}, and \ref{results on kqapro}, FlexKBQA trained using synthetic data only demonstrates comparable performance when compared to previous in-context learning-based methods.
The experimental results clearly illustrate the significant contribution of EGST to the model's performance. It resulted in a substantial increase of 10.3 and 7.1 in F1 score on GrailQA and WebQSP, respectively, and a notable 10.2 increase in accuracy on KQA Pro. These findings provide strong evidence of the effectiveness of EGST in mitigating the impact of distribution shift.
Inherent Reasoning can be regarded as an effective and flexible enhancement method that leverages the inherent knowledge of large models. On the GrailQA and WebQSP datasets, Inherent Reasoning leads to a F1 score improvement of 1.4 (EM score remains unchanged as it focuses solely on the consistency of structural queries) and 2.4, respectively. Remarkably, on the KQA Pro dataset, Inherent Reasoning achieves a accuracy increase of 13.5. This is because the absence of the entity linking stage results in numerous SPARQL execution errors. However, LLM was able to provide accurate answers directly in such cases.

\begin{figure}[htbp]
\centering
\includegraphics[width=0.675\columnwidth]{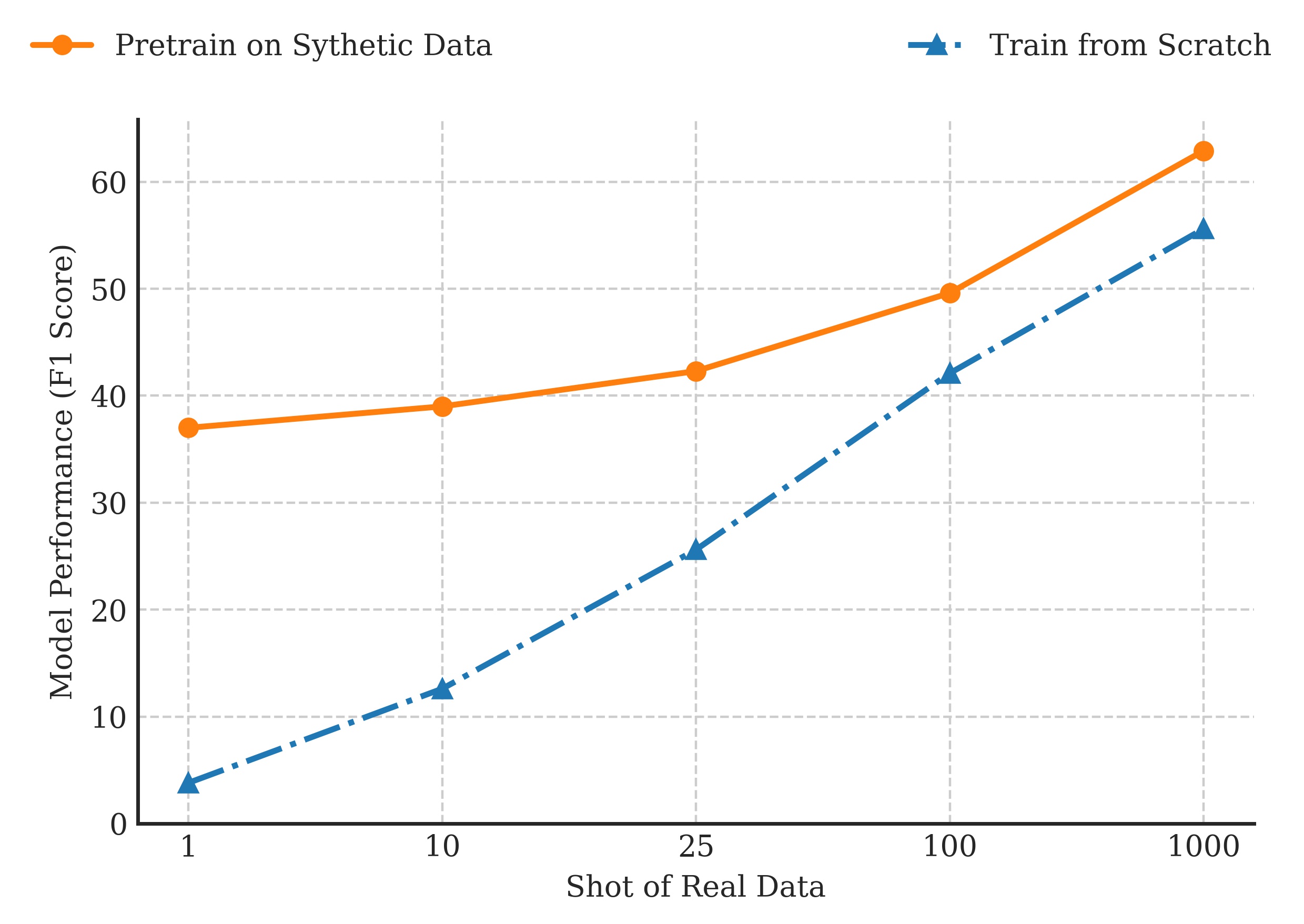} 
\caption{Results beyond few-shot setting. FlexKBQA consistently performs better with more annotated data.}
\label{fig3}
\end{figure}


In conclusion, FlexKBQA achieves superior performance compared to previous few-shot KBQA systems. Moreover, on certain datasets, it demonstrates competitive results when compared to supervised models. 
We claim that the superiority  two main reasons. Firstly, it benefits from a large amount of synthetic data compared to in-context learning, leading to better convergence with respect to data distribution. Secondly, the innovative EGST and IR approaches equip FlexKBQA with the capability to harness unlabeled real user questions and leverage the inherent prowess of LLMs. These unique features distinguish FlexKBQA from other methods.
\subsection{Beyond Few-Shot KBQA}
In this section, we delve into its efficacy beyond the few-shot scenario. As demonstrated in Figure \ref{fig3}, with an increase in the number of samples, the model pretrained on our synthetic data consistently performs better. Even when there are 1000 real samples available, it still maintains an 8-point advantage over models trained solely on real samples. This observation underscores that our approach is not only applicable to few-shot scenarios but can also serve as a valuable data augmentation technique.

\begin{figure}[t]
\centering
\includegraphics[width=0.75\columnwidth]{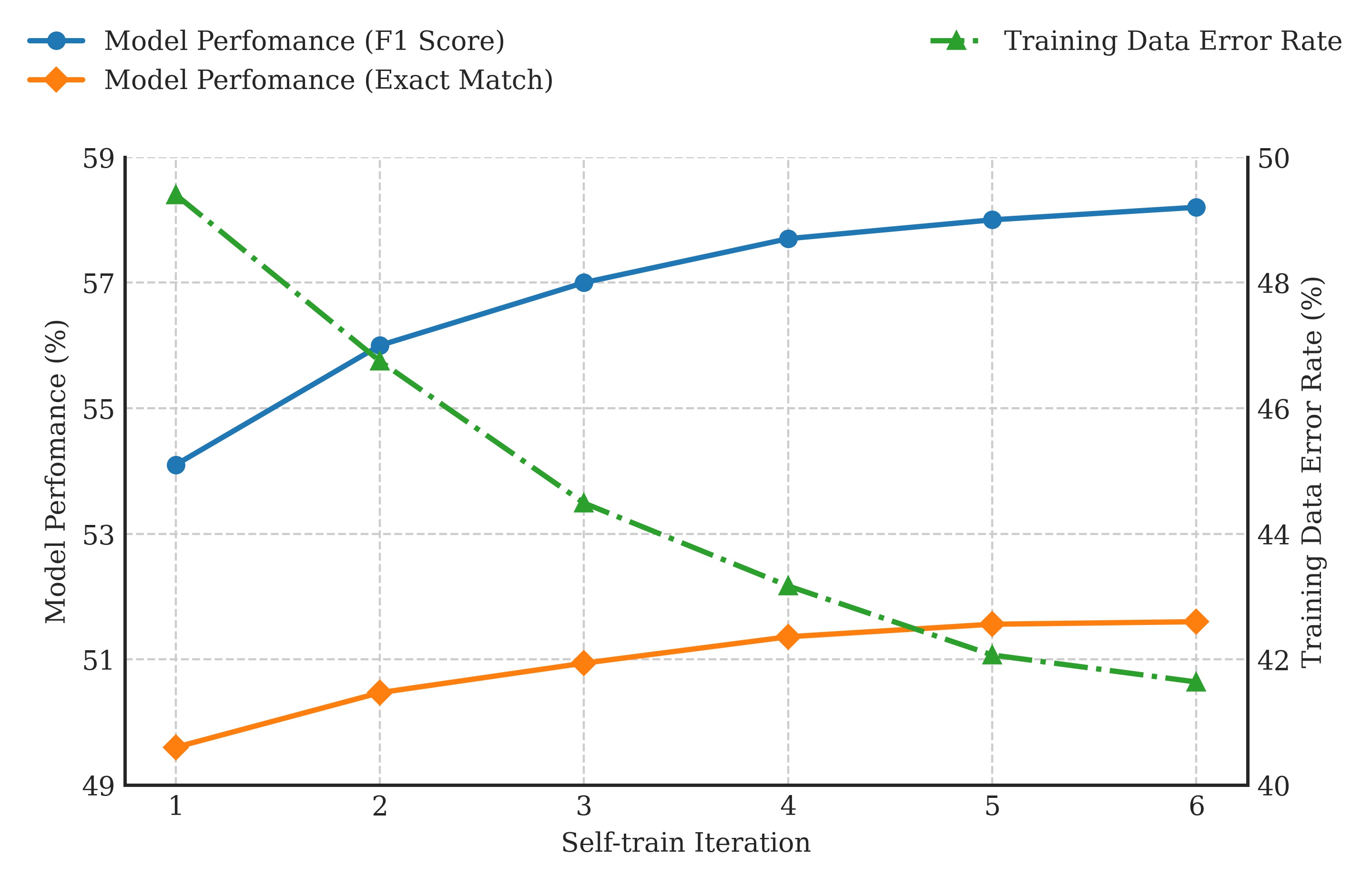} 
\caption{Variation of model performances and the error rate of pseudo-labeled programs with EGST Iterations.}
\label{fig4}
\end{figure}
\subsection{Ablation and Case Studies}
In Figure \ref{fig4}, we delve into an ablation study to gauge the impact of EGST. We can observe that with each iteration, the model's performance steadily enhances, eventually converging around the 6th epoch. The underlying reason, illustrated by the green line, is the diminishing error rate of pseudo programs. We also conduct ablation experiments regarding synthetic data and the number of in-context learning samples in Appendix F and G.

To better showcase the advantages of FlexKBQA, we conduct a comparison between FlexKBQA and Pangu, a state-of-the-art model. In Table \ref{table5}, Question \uppercase\expandafter{\romannumeral1}  illustrates a non-i.i.d scenario, where the relation involved in the question is not present in in-context examples. Consequently, Pangu struggles to answer correctly. However, due to FlexKBQA's inclusion of the relevant relation in its synthetic data, it manages to generate the correct answer. In an ideal situation where the synthetic data coverage is extensive enough, non-i.i.d scenarios like this could be eradicated. Question \uppercase\expandafter{\romannumeral2} demonstrates the impact of EGST. By utilizing information from a correctly pseudo-labeled user question, the model can successfully answer similar but ambiguous or more complex questions in the test set.

\section{Conclusion and Future Work}
In conclusion, this paper presented FlexKBQA, a framework that harnesses the power of LLMs as program translators and their inherent reasoning ability for KBQA tasks. Extensive experiments on diverse datasets showcase the effectiveness of FlexKBQA under the few-shot setting, surpassing all baseline methods and even approaching the performance of supervised models. Furthermore, we are pioneers in exploring zero-shot KBQA setting. FlexKBQA represents a significant advancement in the integration of large and lightweight models, offering three-fold flexibilities: data annotation, deployment, and being domain agnostic. Future research can explore the broader applications of this framework in more natural language understanding and reasoning tasks.

\bibliography{aaai24}

\clearpage

\appendix
\section{Appendix}
In this supplementary material, we furnish the additional details that were left out in the main text. The appendices are organized as follows:
\begin{itemize}
    \item Appendix A: Automatic Program Sampling
    \item Appendix B: Example Prompts and Inherent Reasoning
    \item Appendix C: Execution-Guided Filtering
    \item Appendix D: Underlying models and Baselines
    \item Appendix E: Implementation Details
    \item Appendix F: Synthetic vs. Real Data
    \item Appendix G: Number of In-context Examples
    \item Appendix H: Progressive Execution-guided Self-training
\end{itemize}

 \section{A. Automatic Program Sampling}
\subsection{A.1 Template Collection}
We replace the entities and relations within programs with variables (e.g., $ent0, rel0, ent1$) to get program templates. 
Different program templates capture the fundamental logic of various question types, so it's essential to have a diverse range of templates. In practical settings, people can customize and design their own templates based on specific business scenarios.

In the experiments, we construct templates using the programs included in the few-shot annotated samples. Here is an example of a SPARQL template:

\begin{listing}[ht]
$ \textbf{SELECT}\ (\ ?x0\ AS\ ?value\ )  \ \textbf{WHERE}$
    
$ \quad \{\ ?x0\ \textbf{:type.object.type}\ ?ent1\ . $

$ \quad \textbf{VALUES} \ ?x1\ \{\ ?ent0\ \}$

$ \quad?x0\ ?rel0\ ?x1\ .$

$\quad \textbf{FILTER}\ (\ ?x0\ !=\ ?x1\  )\ \} $
\end{listing}
For S-expressions, we first convert them into SPARQL queries\footnote{https://github.com/dki-lab/ArcaneQA/issues/3} and then follow the same procedure.

\subsection{A.2 Step-wise Program Grounding}
We directly treat the variables in a template as query objects. The aforementioned example can be rephrased as the following query:
\begin{listing}[ht]
$ \textbf{SELECT}\ ?rel0\  ?ent0\ ?ent1  \ \textbf{WHERE}$
    
$ \quad \{\ ?x0\ \textbf{:type.object.type}\ ?ent1\ . $

$ \quad \textbf{VALUES} \ ?x1\ \{\ ?ent0\ \}$

$ \quad?x0\ ?rel0\ ?x1\ .$

$\quad \textbf{FILTER}\ (\ ?x0\ !=\ ?x1\  )\ \} $
\end{listing}

 Then we introduce a ``step-wise grounding" approach, where we iteratively determine the values of variables. 
 In the given example, a plausible order for establishing variable values could be $ent0 \rightarrow rel0 \rightarrow ent1$. Once these variables are grounded to specific values, we can populate them into the original template to form an executable program.
\section{B. Example Prompts and Inherent Reasoning}
We show examples of prompts in Listing 1, 2, and 3 for GrailQA, WebQSP and KQA Pro, respectively. The prompt design is straightforward. More efficient prompting techniques like batch prompting \cite{cheng2023batch} remain to be explored.

For Inherent Reasoning, we also employ the in-context learning approach, but the prompts are directly constructed from question-answer pairs, which can be easily collected from the web. For convenience, we utilize question-answer pairs from the dataset.
\section{C. Execution-Guided Filtering}

For semantic filtering, we employ an off-the-shelf pre-trained sentence-transformer named all-MiniLM-L6-v2 {\cite{reimers2019sentencebert}}, which maps sentences to a 384-dimensional dense vector space. 
We use the pre-trained model\footnote{\url{https://huggingface.co/sentence-transformers/all-MiniLM-L6-v2}} to calculate the semantic textual similarity between natural language questions and the relations in predicted programs. The semantic similarity is caculated as the cosine similarity of the embedding vectors, and we average the similarity scores if there are multiple relations in one predicted program. Empirically, we set the threshold of 0.2 to filter out those data pairs with similarity below the threshold. The semantic filtering technique can effectively filter out data pairs whose predicted relations are irrelevant to the natural language questions.

For inherent filtering, once the surface name of the predicted answer entity derived from the program appear in the direct response provided by the large language model, we consider this pseudo-label program valid. This approach can enhance the purity of pseudo-labeled training data.

Additionally, We also remove data pairs where the surface names of retrieved entities cannot be obtained through the sparql endpoint to Freebase, as these entities are quite uncommon and usually lead to erroneous results.

\section{D. Underlying models and Baselines}
\subsection{D.1 Underlying models}

We select advanced models that have demonstrated outstanding performance on KBQA datasets. 
Specifically, for GrailQA and WebQSP, we choose the RnG-KBQA \cite{ye2022rng} model. For KQA Pro, we select the BART-SPARQL \cite{cao2022kqa} model. They also demonstrate good reproducibility.
 \newline
\textbf{RnG-KBQA} follows a ranking-then-generate approach. It begins with searching for a collection of candidate programs on the graph, using the topic entity mentioned in the question. These candidates are then ranked by similarity scores outputted by a ranker model. Finally, a pretrained generation model, serving as the generator consumes both the question and the top-k candidate programs to generate the final answer. In experiments, we initialize the ranker with the BERT-base-uncased model, and the generator with T5-base.
\textbf{BART-SPARQL}, in contrast, is an end-to-end generation model that directly produces the corresponding SPARQL query given a question. It is worth noting that the pre-trained BART model is forced to have the capability to memorize the relations and entities present in the KB. As a result, the BART-SPARQL model is more sensitive to unseen questions compared to RnG-KBQA. We also use the BART-base for this model.

\subsection{D.2 Baselines}
\textbf{Pangu} is a recently emerged state-of-the-art model. It
proposes using LLMs for discrimination rather than generation in grounded language understanding, combining symbolic search with neural scoring in a novel way. It incrementally constructs plans in a step-wise fashion to handle large search spaces. At each step, it extends existing plans into new valid candidates based on the environment. This allows it to guarantee the grammaticality and faithfulness of the proposed plans. \newline 
\textbf{KB-BINDER} enables few-shot learning for KBQA using LLMs through two key stages: Draft Generation, where given a question, an LLM generates a preliminary ``draft" logical form by imitating few examples that capture semantic relationships but may contain mistakes; and Knowledge Base Binding, where entities and relations in the draft are grounded to the target KB using string matching and similarity search to produce refined, executable candidate logical forms. \newline
\textbf{Fine-Tuning} means that we train the model only on the few-shot data (e.g., 100 shots). This baseline is designed to demonstrate how existing models' performance significantly decreases when faced with insufficient data, thereby highlighting the superiority of our framework. \newline
\textbf{LLM-ICL} is an in-context learning-based baseline we implement for KQA Pro. As there are no experimental results of Pangu and KB-BINDER on KQA Pro, we use LLM-ICL as an alternative. Since KQA Pro models do not include an entity linking stage, LLM-ICL directly generates SPARQL queries without further grounding stage, ensuring a fair comparison.

We also presented the performance of several supervised models (e.g., ReTraCk \cite{chen2021retrack}, DecAF \cite{yu2023decaf}) to demonstrate that FlexKBQA maintains usability while offering flexibility.
\section{E. Implementation Details}
On GrailQA and WebQSP datasets, we use the underlying model RnG-KBQA \cite{ye2022rng}. Empirically, we adopted the non-bootstrap ranking strategy in the RnG-KBQA. All other settings remain consistent with the original setting. Experiments can be conducted with a single GeForce RTX 3090 graphics card.

 For the entity linking, we adopt the same setting as our baseline models. On GrailQA, we use the entity linking results from TIARA \cite{shu2022tiara}. On WebQSP, we adopted the entity linking results from ELQ \cite{li2020efficient}.

 For the second stage of inherent reasoning augmentation, we resort to using the responses from LLM as the final answers for the initially unanswerable questions. Specifically, we perform the Named Entity Recognition (NER) techniques on the textual responses and take the recognized entities as the final answers. The NER system we use is also consistent with the NER system in RnG-KBQA.

\section{F. Synthetic vs. Real Data}
To better demonstrate the gap between the synthetic data generated by our framework and the real data, as well as to explore the reasons behind this discrepancy, we conducted this supplementary experiment. We compare the model performance when trained using diverse data sources:
\begin{itemize}
\item \textbf{Synthetic Data}: we produce  around 1,000 synthetic data using our automatic program sampling and low-resource program translation method. We use those synthetic data to train the RnG-KBQA ranker model and evaluate the model performance on the WebQSP test set.

    \item \textbf{Real Data}: we randomly sample the same amount of training data directly from the original WebQSP dataset.
    
    \item \textbf{Real Program, Synthetic Question}: we leverage the real programs of training samples in the “real data” setting, and use the low-resource program translation to generate the natural language questions. 
\end{itemize}
The model performances under these three settings are demonstrated in Table \ref{Discrepancy between Synthetic and Real Data}. The performance of the model trained by synthetic data is significantly worse than that trained by real data. This is largely due to the fact that when generating synthetic data, the programs we sample are difficult to form similar distribution as the test set, such as containing consistent relations. This scenario is more significant on the strong \textit{i.i.d.} WebQSP dataset.
In the ``real program, synthetic question" setting, the model performance is just slightly weaker than the model trained by real data. This implies that LLMs have a strong ability to translate symbolic language into natural language.





\begin{table}[t]
  \centering
    \begingroup
  \begin{tabular}{cccc} \toprule
  \textbf{Setting} &  \ \textbf{EM} & \ \textbf{F1}  \\ \midrule
Real Data  & \textbf{61.4}  & \textbf{68.2}  \\ 
\midrule

Sythetic Data  & 32.7 & 35.5 \\ 
\midrule

Real Program, Synthetic Question   & 50.9 & 56.3 \\ 

 \bottomrule
  \end{tabular}
  \caption{Discrepancy between Synthetic and Real Data}
  \label{Discrepancy between Synthetic and Real Data}
  \endgroup
\end{table}

\section{G. Number of In-context Examples}
We conduct an ablation study concerning the number of in-context learning examples in our framework. We conduct the experiment on the GrailQA dataset and evaluate the model performance using the GrailQA dev set and test set. To reduce interference and focus on the impact of the number of in-context examples on data generation, we do not perform the inherent reasoning augmentation in this ablation experiment. The F1 scores are shown in Table \ref{Number of In-context Examples}. We can observe that the number of examples has little impact on the final performance of the model, as the knowledge about program languages (e.g. SPARQL, S-expression) in the examples has already been largely incorporated into the parameters of the LLMs.

\begin{table}[tb]
  \centering
  \scalebox{0.9}{
  \begin{tabular}{c|l|c|c}
  \toprule
 \textbf{\# Exmaples} &\textbf{Model} & \textbf{F1(dev)} &\textbf{F1(test)}\\
\midrule
\multirow{3}{*}{10}
&   FlexKBQA &  \\
&   \quad \textit{-w/o} IR & 70.04 & 66.61 \\
&   \quad \textit{-w/o} EGST & 61.78 & 57.70\\
\midrule 
\multirow{3}{*}{25}
&   FlexKBQA &  \\
&   \quad \textit{-w/o} IR & 70.23 & 67.19 \\
&   \quad \textit{-w/o} EGST & 62.10 & 57.74\\
\midrule
\multirow{3}{*}{60}
&   FlexKBQA &  \\
&   \quad \textit{-w/o} IR & 70.29 & 67.20 \\
&   \quad \textit{-w/o} EGST & 62.54 & 58.71\\

  \bottomrule
  \end{tabular}}
\caption{Number of In-context Examples}
  \label{Number of In-context Examples}
\end{table}

\section{H. Progressive Execution-Guided Self-Training}
During the execution-guided filtering phase, we have multiple filtering strategies available. Exploring effective ways to utilize them is still worth investigating.
Here we also explore a strategy called Progressive Execution-Guided Self-Training (\textbf{PEGST}), a slightly distinct variant of EGST. In PEGST, we employed different filtering strategies in different stages:
\begin{itemize}
    \item \textbf{Stage \uppercase\expandafter{\romannumeral1}}: Error Filtering
    \item \textbf{Stage \uppercase\expandafter{\romannumeral2}}: Semantic Filtering
    \item \textbf{Stage \uppercase\expandafter{\romannumeral3}}: We remove data pairs where the surface names of retrieved entities cannot be obtained through the SPARQL endpoint to Freebase.
\end{itemize}

\begin{figure}[t]
\centering
\includegraphics[width=0.9\columnwidth]{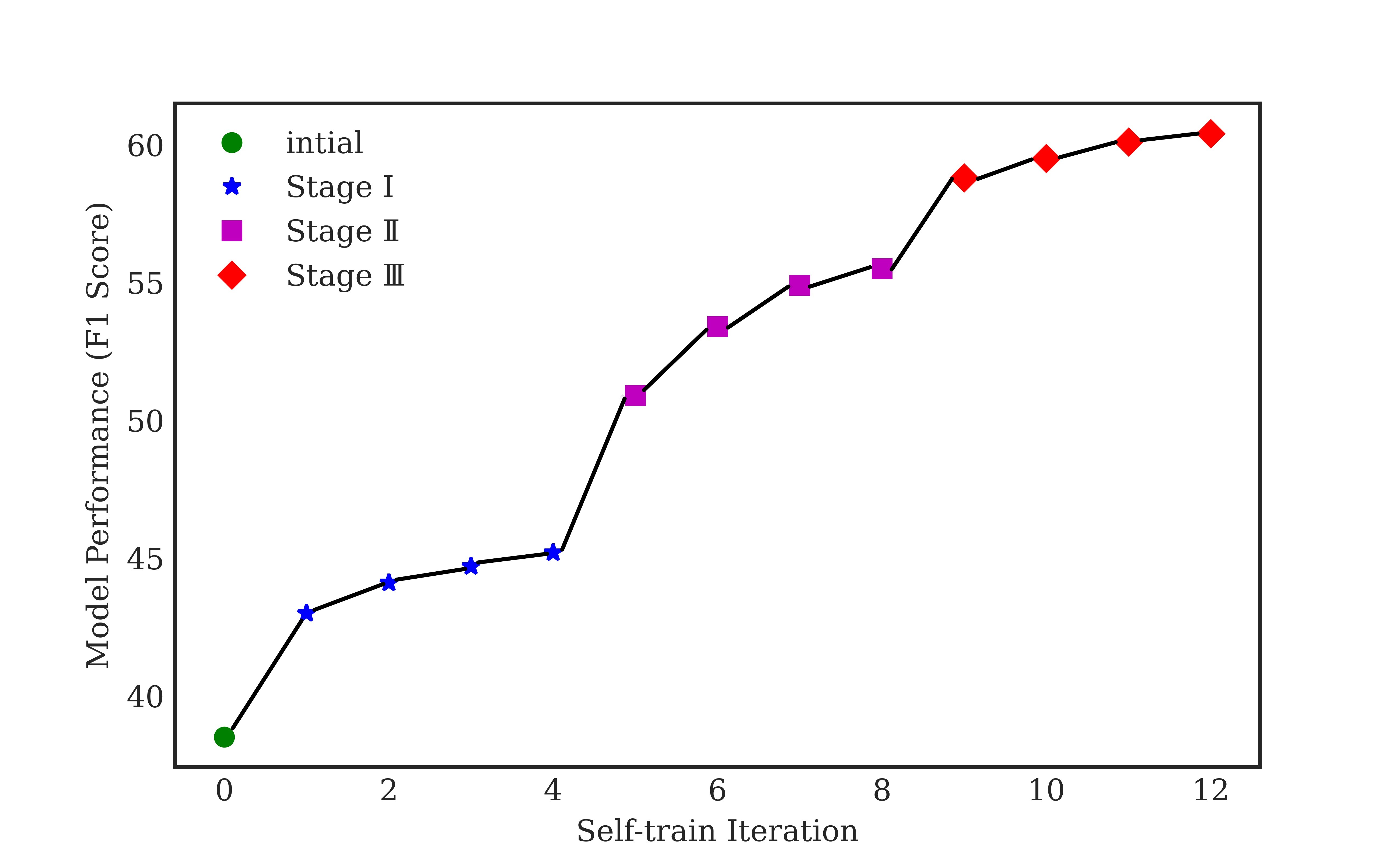} 
\caption{Model improvement in the PEGST process}
\label{fig5}
\end{figure}

We conduct an experiment of PEGST on the WebQSP dataset with other settings remaining the same. From the increasing trend of the F1 score in Figure \ref{fig5}, we can notice that in PEGST, a substantial enhancement in the model's capabilities can be achieved once a criterion for filtering data is altered. However, within a single stage, the increase in the model's capacity is relatively minor. PEGST can be considered as a process of injecting diverse knowledge into the model in distinct phases. In experiments, PEGST can achieve a slight advantage over EGST, but it requires more self-training iterations and a longer period to execute. Consequently, we opted for employing the EGST approach in the main body of our study. 

\clearpage
\quad \textbf{Listing 1. Example Prompt for GrailQA}
\newline

\lstset{backgroundcolor=\color[RGB]{245,245,244},
    linewidth=17cm,
    numbers=none,
    basicstyle=\ttfamily\small
}\begin{lstlisting}
### Convert the s-expressions to natural language questions.

# s-expression:(AND medicine.routed_drug (JOIN medicine.routed_drug.marketed_formulations Oxybutynin chloride 5 extended release film coated tablet))
# question:oxybutynin chloride 5 extended release film coated tablet is the ingredients of what routed drug?

# s-expression:(ARGMAX food.food food.food.energy)
# question:when it comes to the food that has the most energy per 100g what is the name of it?

# s-expression:(AND music.genre (JOIN (R music.genre.parent_genre) (JOIN music.genre.albums confessions tour)))
# question:the albums confessions tour is part of what parent genre of a musical genre?

# s-expression:(AND architecture.architect (JOIN architecture.architect.architectural_style (JOIN (R architecture.architect.architectural_style) Josef Fanta)))
# question:which architect has a similar architectural style to josef fanta?

# s-expression:(AND architecture.building (lt architecture.building.floors 9^^http://www.w3.org/2001/XMLSchema#integer))
# question:which building has less than 9 floors?

# s-expression:(AND comic_books.comic_book_series (JOIN (R comic_books.comic_book_genre.comic_book_series_in_this_genre) (JOIN (R comic_books.comic_book_story.genre) Case of the Namesake Murders)))
# question:case of the namesake murders is the same genre as what comic book series?

# s-expression:(AND film.director (JOIN (R media_common.quotation.author) It is an open question whether any behavior based on fear of eternal punishment can be regarded as ethical or should be regarded as merely cowardly.))
# question:what is the name of the author who wrote it is an open question whether any behavior based on fear of eternal punishment can be regarded as ethical or should be regarded as merely cowardly.?

# s-expression:(AND meteorology.beaufort_wind_force (ge meteorology.beaufort_wind_force.wave_height 7.0^^http://www.w3.org/2001/XMLSchema#float))
# question:for waves higher than 7.0 what is the beaufort window force?

# s-expression:(AND book.short_story (JOIN book.short_story.characters (JOIN book.book_character.appears_in_stories Doing Clarence a Bit of Good)))
# question:what short story has a character who also is in doing clarence a bit of good?

# s-expression:(AND education.school_category (AND (JOIN (R education.educational_institution.school_type) Chiang Kai Shek College) (JOIN education.school_category.schools_of_this_kind Sacred Heart High School (Roseville, Michigan))))
# question:chiang kai shek college and sacred heart high school (roseville, michigan) are in what category of school?

# s-expression:(ARGMIN food.bottled_water food.bottled_water.nitrate_content)
# question:the bottled water that has the least amount of nitrates?

......

# s-expression:
\end{lstlisting}

\clearpage
\quad \textbf{Listing 2. Example Prompt for WebQSP}
\newline

\lstset{
    backgroundcolor=\color[RGB]{245,245,244},
    linewidth=17cm,
    numbers=none,
    basicstyle=\ttfamily\small
}\begin{lstlisting}
### Convert the s-expressions to natural language questions.

# s-expression:(JOIN (R location.location.containedby) Auburn University)
# question:where is university of auburn?

# s-expression:(JOIN (R government.political_party_tenure.politician) (JOIN (R government.political_party.politicians_in_this_party) New Democratic Party))
# question:who founded the new democratic party?

# s-expression:(ARGMAX (JOIN (R location.location.contains) China) topic_server.population_number)
# question:where do most chinese live?
	
# s-expression:(JOIN (R film.performance.actor) (AND (JOIN film.performance.character Dorothy Gale) (JOIN (R film.film.starring) The Wizard of Oz)))
# question:who played dorothy in the film wizard of oz?

# s-expression:(AND (JOIN common.topic.notable_types College/University) (JOIN (R education.education.institution) (JOIN (R people.person.education) Jerry Rice)))
# question:what college did jerry rice attend?

# s-expression:(AND (JOIN common.topic.notable_types City/Town/Village) (JOIN (R location.location.contains) Oakland County))
# question:what cities are in oakland county michigan?

# s-expression:(JOIN (R people.marriage.spouse) (AND (JOIN people.marriage.time_macro 2015^^http://www.w3.org/2001/XMLSchema#date) (AND (JOIN people.marriage.type_of_union Marriage) (JOIN (R people.person.spouse_s) Jane Krakowski))))
# question:who is married to jane krakowski?

# s-expression:(AND (JOIN people.person.gender Female) (JOIN (R people.marriage.spouse) (ARGMAX (JOIN (R people.person.spouse_s) Robert Downey Jr.) people.marriage.from)))
# question:who is robert downey jr wife?

# s-expression:(JOIN (R location.religion_percentage.religion) (ARGMAX (JOIN (R location.statistical_region.religions) United States of America) location.religion_percentage.percentage))
# question:what is the most practiced religion in the united states?

# s-expression:(ARGMAX (AND (JOIN sports.sports_championship_event.champion Dallas Cowboys) (JOIN (R sports.sports_team.championships) Dallas Cowboys)) time.event.end_date)
# question:when was the last dallas cowboys super bowl win?

# s-expression:(ARGMAX (JOIN (R film.performance.film) (JOIN (R film.actor.film) Brittany Murphy)) film.film.initial_release_date)
# question:what is the last movie brittany murphy made?

# s-expression:(AND (JOIN book.written_work.subjects Evolution) (AND (JOIN common.topic.notable_types Book) (JOIN (R book.author.works_written) Charles Darwin)))
# question:what book did charles darwin write on evolution?

......

# s-expression:
\end{lstlisting}

\clearpage
\quad \textbf{Listing 3. Example Prompt for KQA Pro}
\newline

\lstset{
    backgroundcolor=\color[RGB]{245,245,244},
    linewidth=17cm,
    numbers=none,
    basicstyle=\ttfamily\small
}\begin{lstlisting}
### Convert the sparqls to natural language questions.

# sparql:SELECT DISTINCT ?e WHERE { ?e <pred:instance_of> ?c . ?c <pred:name> "town" . ?e <TOID> ?pv . ?pv <pred:value> "4000000074573917" . ?e <OS_grid_reference> ?pv_1 . ?pv_1 <pred:value> "SP8778" .  }
# question:Which town has a TOID of 4000000074573917 and has an OS grid reference of SP8778?

# sparql:SELECT DISTINCT ?pv WHERE { ?e <pred:instance_of> ?c . ?c <pred:name> "human" . ?e <ISNI> ?pv_1 . ?pv_1 <pred:value> "0000 0001 2136 4821" . ?e <date_of_birth> ?pv .  }
# question:When was the person with ISNI 0000 0001 2136 4821 born?

# sparql:ASK { ?e <pred:name> "Eve Myles" . ?e <official_website> ?pv . ?pv <pred:value> "http://www.cheechandchong.com" .  }
# question:Is http://www.cheechandchong.com Eve Myles's official website?

# sparql:SELECT DISTINCT ?p WHERE { ?e_1 <pred:name> "alternative rock" . ?e_2 <pred:name> "Greg Graffin" . ?e_1 ?p ?e_2 .  }
# question:What has alternative rock in common with Greg Graffin?

# sparql:ASK { ?e <pred:instance_of> ?c . ?c <pred:name> "human" . ?e <ISNI> ?pv_1 . ?pv_1 <pred:value> "0000 0001 0893 552X" . ?e <name_in_native_language> ?pv . ?pv <pred:value> "Elias Koteas" .  }
# question:Is Elias Koteas the name in native language of the person with ISNI 0000 0001 0893 552X?

# sparql:SELECT DISTINCT ?qpv WHERE { ?e_1 <pred:name> "Gregg Allman" . ?e_2 <pred:name> "Cher" . ?e_2 <birth_name> ?pv . ?pv <pred:value> "Cherilyn Sarkisian" . ?e_1 <spouse> ?e_2 . [ <pred:fact_h> ?e_1 ; <pred:fact_r> <spouse> ; <pred:fact_t> ?e_2 ] <end_time> ?qpv .  }
# question:When did Gregg Allman stop being the spouse of Cher (the one whose birth name is Cherilyn Sarkisian)?

# sparql:SELECT (COUNT(DISTINCT ?e) AS ?count) WHERE { ?e <pred:instance_of> ?c . ?c <pred:name> "town" . ?e <postal_code> ?pv . ?pv <pred:value> "VLT" . ?e <area> ?pv_1 . ?pv_1 <pred:unit> "square mile" . ?pv_1 <pred:value> ?v . FILTER ( ?v < "530"^^xsd:double ) .  }
# question:How many towns' postal code is VLT and area is less than 530 square miles?

# sparql:SELECT (COUNT(DISTINCT ?e) AS ?count) WHERE { ?e <pred:instance_of> ?c . ?c <pred:name> "sovereign state" . ?e <population> ?pv . ?pv <pred:unit> "1" . ?pv <pred:value> ?v . FILTER ( ?v != "7600000000"^^xsd:double ) .  }
# question:What number of sovereign states have a population not equal to 7600000000?

# sparql:SELECT DISTINCT ?pv WHERE { ?e <pred:name> "63rd Golden Globe Awards" . ?e <edition_number> ?pv .  }
# question:What is the edition number of the 63rd Golden Globe Awards?

# sparql:ASK { ?e <pred:instance_of> ?c . ?c <pred:name> "film" . ?e <official_website> ?pv_1 . ?pv_1 <pred:value> "http://www.lovelybones.com" . ?e <duration> ?pv . ?pv <pred:unit> "minute" . ?pv <pred:value> ?v . FILTER ( ?v > "60"^^xsd:double ) .  }
# question:Does the film, whose official website is http://www.lovelybones.com, have a duration greater than 60 minutes?

......

# sparql:
\end{lstlisting}

\end{document}